\documentclass[letterpaper, 10 pt, conference]{ieeeconf}  % Comment this line out if you need a4paper

\IEEEoverridecommandlockouts                              % This command is only needed if 
\title{\LARGE \bf
DropoutDAgger: A Bayesian Approach to Safe Imitation Learning
}

\author{Kunal Menda, Katherine Driggs-Campbell, and Mykel J. Kochenderfer
\thanks{
	This material is based upon work supported by SAIC Motor.}% <-this % stops a space
\thanks{K. Menda, K. Driggs-Campbell, and M.J. Kochenderfer are with the Aeronautics and Astronautics Department at Stanford University, Stanford, CA USA (e-mail: \{kmenda,krdc,mykel\}@stanford.edu).}%
}

\usepackage[letterpaper, portrait, margin=1in]{geometry}

\usepackage{cite}
\usepackage{array}
\usepackage{graphics} % for pdf, bitmapped graphics files
\usepackage{subfig}
\usepackage{multirow}

\usepackage{suffix}
\usepackage{xstring}
\usepackage{xparse}
\usepackage{expl3}

\usepackage{url}
\usepackage{dsfont}
\usepackage{algorithm}
\usepackage{algorithmicx}
\usepackage[noend]{algpseudocode}
\usepackage{graphicx,caption}
\usepackage{amsmath,amssymb,stmaryrd,mathtools}
\usepackage[keeplastbox]{flushend}
\usepackage{algorithm}
\usepackage{cite}
\usepackage{hyperref}
\usepackage{cleveref}
\usepackage{acronym}
\usepackage{color,xcolor}
\usepackage{verbatim}
\usepackage{booktabs}
\usepackage{siunitx}
\graphicspath{{./figures/}}

\newcommand{\ignore}[1]{}
\DeclareFontFamily{U}{MnSymbolC}{}
\DeclareSymbolFont{MnSyC}{U}{MnSymbolC}{m}{n}
\DeclareFontShape{U}{MnSymbolC}{m}{n}{
    <-6>  MnSymbolC5
   <6-7>  MnSymbolC6
   <7-8>  MnSymbolC7
   <8-9>  MnSymbolC8
   <9-10> MnSymbolC9
  <10-12> MnSymbolC10
  <12->   MnSymbolC12%
}{}
\DeclareMathSymbol{\powerset}{\mathord}{MnSyC}{180}

\definecolor{orange}{rgb}{1, .36, .08}
\definecolor{darkmagenta}{rgb}{0.698,0,0.698}

\definecolor{vg_edit_color}{rgb}{0, 0.0, 1.0}

\usepackage{todonotes}
\usepackage{soul}
\definecolor{smoothgreen}{rgb}{0.7,1,0.7}
\sethlcolor{smoothgreen}

\RequirePackage{luatex85}
\usepackage{pgfplots}
\pgfplotsset{compat=newest}
\pgfplotsset{every axis legend/.append style={%
cells={anchor=west}}
}
\usepgfplotslibrary{polar}
\usetikzlibrary{arrows}
\tikzset{>=stealth'}

\definecolor{C1}{rgb}{0.0, 0.447, 0.741}
\definecolor{C1_light}{rgb}{0.0, 0.6032388663967612, 1.0}
\definecolor{C2}{rgb}{0.85, 0.325, 0.098}
\definecolor{C3}{rgb}{0.929, 0.694, 0.125}
\definecolor{C4}{rgb}{0.494, 0.184, 0.556}
\definecolor{C5}{rgb}{0.466, 0.674, 0.188}
\definecolor{C6}{rgb}{0.301, 0.745, 0.933}
\definecolor{C7}{rgb}{0.635, 0.078, 0.184}

\usepgfplotslibrary{groupplots}

\usetikzlibrary{shapes.geometric, arrows}

\tikzstyle{startstop} = [rectangle, rounded corners, minimum width=2cm, minimum height=1cm,text centered, draw=black, fill=none]
\tikzstyle{arrow} = [thick,->,>=stealth]

\begin{document}

\maketitle
\thispagestyle{empty}
\pagestyle{empty}

%%%%%%%%%%%%%%%%%%%%%%%%%%%%%%%%%%%%%%%%%%%%%%%%%%%%%%%%%%%%%%%%%%%%%%%%%%%%%%%%
\begin{abstract}
While imitation learning is becoming common practice in robotics, this approach often suffers from data mismatch and compounding errors.
DAgger is an iterative algorithm that addresses these issues by continually aggregating training data from both the expert and novice policies, but does not consider the impact of safety. %, which is crucial when failures have harsh consequences.
We present a probabilistic extension to DAgger, which uses the distribution over actions provided by the novice policy, for a given observation. 
Our method, which we call DropoutDAgger, uses dropout to train the novice as a Bayesian neural network that provides insight to its confidence.
Using the distribution over the novice's actions, we estimate a probabilistic measure of safety with respect to the expert action, tuned to balance exploration and exploitation.
The utility of this approach is evaluated on the MuJoCo HalfCheetah and in a simple driving experiment, demonstrating improved performance and safety compared to other DAgger variants and classic imitation learning.
\end{abstract}

%%%%%%%%%%%%%%%%%%%%%%%%%%%%%%%%%%%%%%%%%%%%%%%%%%%%%%%%%%%%%%%%%%%%%%%%%%%%%%%%
\section{INTRODUCTION}

Recently, there have been many advances in robotics driven by breakthroughs in deep imitation learning \cite{Kober2010,ARGALL2009469}.
Yet, to be truly intelligent, such systems must have the ability to explore their state-space in a safe way \cite{amodei2016concrete}.
One method to guide exploration is to learn from expert demonstrations~\cite{price2003accelerating,schaal1997learning}.
In contrast with reinforcement learning, where an explicit reward function must be defined, imitation learning guides exploration through expert supervision, allowing our robot policy to effectively learn directly from experiences \cite{ARGALL2009469}.

However, such supervised approaches are often suboptimal or fail when the policy that is being trained (referred to as the novice policy) encounters new situations or enters a state that is poorly represented in the dataset provided by the expert \cite{daume2009search,ross2010efficient}.
While failures may be insignificant in simulation, safe learning is of the utmost importance when acting in the real-world \cite{amodei2016concrete}.

Methods for guided policy search in imitation learning settings have been developed~\cite{levine2013guided}.
An example of these approaches is \textsc{DAgger}, which improves the data represented in the training dataset by continually aggregating new data from both the expert and novice policies.
\textsc{DAgger} has many desirable properties, including online functionality and theoretical guarantees.
This approach, however, does not provide safety guarantees.
Recent work extended \textsc{DAgger} to address some inherent drawbacks~\cite{kim2013maximum,laskey2016shiv}.
In particular, \textsc{SafeDAgger} augments \textsc{DAgger} with a decision rule policy to provide safe exploration with minimal influence from the expert~\cite{Zhang2016}.

This paper augments \textsc{DAgger} by extending the approach to a probabilistic domain.
We build upon the \textsc{SafeDAgger} idea of safety by considering the distribution over the novice's actions.
This approach allows us to glean some insight into the deep policy through a notion of confidence, in addition to safety bounds.

We demonstrate how our method out-performs existing algorithms in classical imitation learning settings. 
This paper presents three key contributions:
\begin{enumerate}
	\item We develop a probabilistic notion of safety to balance exploration and exploitation;
	\item We present \textsc{DropoutDAgger}, a Bayesian extension to \textsc{DAgger}; and 
	\item We demonstrate the utility of this approach with improved performance and safety in imitation learning case studies.
\end{enumerate}

This paper is organized as follows.
\Cref{sec:related_works} provides a brief overview of the underlying principles to be employed in this work.
The methodology behind \textsc{DropoutDAgger} is presented in \Cref{sec:methods}.
The two experimental settings used to validate our approach are described in \Cref{sec:experiments}.
\Cref{sec:conclusion} discusses our findings and outlines future work.

%%%%%

\section{BACKGROUND}
\label{sec:related_works}

This section presents a brief technical overview of \textsc{DAgger}, \textsc{SafeDAgger}, and dropout as applied to Bayesian neural networks.

\subsection{DAgger and SafeDAgger}
The \textsc{DAgger} framework extends traditional supervised learning approaches by simultaneously running both an expert policy that we wish to clone and a novice policy we wish to train \cite{ross2011reduction}.
By constantly aggregating new data samples from the expert policy, the underlying model and reward structure are uncovered.

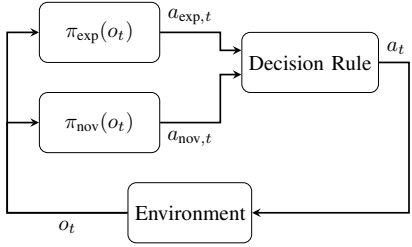
\begin{figure}[!t]
	\centering
	\scalebox{0.8}{% \usetikzlibrary{shapes.geometric, arrows}

% \tikzstyle{startstop} = [rectangle, rounded corners, minimum width=2cm, minimum height=1cm,text centered, draw=black, fill=none]

\begin{tikzpicture}[node distance=1cm]

\node (piexp) [startstop] {$\pi_\text{exp}(o_t)$};
\node (pinov) [startstop, below of=piexp, yshift=-0.5cm] {$\pi_\text{nov}(o_t)$};
\node (decisionsystem) [startstop, right of=piexp, xshift=2.5cm, yshift=-0.5cm] {Decision Rule};
\node (env) [startstop, below of=pinov, xshift=1.5cm, yshift=-0.5cm] {Environment};

\draw [arrow] (piexp.east) -- node[anchor=south] {$a_{\text{exp},t}$} ++(+1.0,0) |- (decisionsystem.170);
\draw [arrow] (pinov.east) -- node[anchor=north] {$a_{\text{nov},t}$} ++(+1.0,0) |- (decisionsystem.190);
\draw [arrow] (decisionsystem.east) |- node[anchor=south west] {$a_t$} ++(+0.5,0) |- (env.east);
\draw [arrow] (env.west) -- node[anchor=north] {$o_t$} ++(-2,0) |- (piexp.west);
\draw [arrow] (env.west) -- ++(-2,0) |- (pinov.west);

% \draw [arrow] (obs.350) -- (beta.186);
% % \draw [arrow] (mux) -- (beta);
% \draw [arrow] (memory.south) |- node[anchor=north] {$m_{t-1}$} ++(-2,-0.5) |-  (beta.170);
% \draw [arrow] (meta.350) -- (selector.190);
% \draw [arrow] (memory.south) |- ++(2,-0.5) |- (selector.170);
% \draw [arrow] (selector.16) -- node[anchor=north] {$m_t$} (lowlev.171);
% \draw [arrow] (selector.16) |- ++(.5, 0) |- ++(0,3.8) -- (memory.east);
% \draw [arrow] (beta.east) -| (selector.north);
% \draw [arrow] (lowlev) -- (action);
% \draw [arrow] (obs.350) -| ++(0.5,-1.9) -- (meta);
% \draw [arrow] (obs.350) -| ++(0.5,-3) -| ++(6.5,0) |- (lowlev.191);

\end{tikzpicture}}
	\caption{\small Flowchart for action selection for DAgger and DAgger variants, where the Decision Rule differs between approaches.
		\label{fig:combinedsystemflowchart}}
\end{figure}

Given some initial training set $\mathcal{D}_0$ generated by the expert policy $\pi_{\text{exp}}$, an initial novice policy $\pi_{\text{nov},0}$ is trained.
Using this initialization, DAgger iteratively collects episodes additional training examples from a mixture of the expert and novice policy. During a given episode, the combined-expert-and-novice system interacts with the environment under the supervision of a decision rule. 
The decision rule decides at every time-step whether the novice's or the expert's choice of action is used to interact with the environment (\Cref{fig:combinedsystemflowchart}). 
The observations received during the episodes of an epoch and the expert's choice of corresponding actions make up a new dataset called $\mathcal{D}_i$. 
The new dataset of training examples is combined with the previous sets: $\mathcal{D} = \mathcal{D} \cup \mathcal{D}_i$, and the novice policy is then re-trained on $\mathcal{D}$. 
The \textsc{DAgger} Algorithm is presented in \Cref{alg:dagger}.

\begin{algorithm}[!t]
	\caption{\textsc{DAgger} \label{alg:dagger}}
	\begin{algorithmic}[1]
		\Procedure{DAgger}{DR($\cdot$)}
		\State Initialize $\mathcal{D} \gets \emptyset $
		\State Initialize $\pi_{\text{nov},i}$
		\For{epoch $i=1:K$}
		\State Sample $T$-step trajectories with $a_t = \text{DR}(o_t)$
		\State Get $\mathcal{D}_i = \left\{s,\pi_\text{exp}(s)\right\}$ of states visited
		\State Aggregate datasets: $\mathcal{D} \gets \mathcal{D}\cap\mathcal{D}_i$
		\State Train $\pi_{\text{nov},i+1}$ on $\mathcal{D}$
		\EndFor
		\EndProcedure
	\end{algorithmic}
\end{algorithm}

By allowing the novice to act, the combined system explores parts of the state-space further from the nominal trajectories of the expert. In querying the expert in these parts of the state-space, the novice is able to learn a more robust policy. However, allowing the novice to always act risks the possibility of encountering an unsafe state, which can be costly in real-world experiments. The vanilla \textsc{DAgger} algorithm and \textsc{SafeDAgger} balance this trade-off by their choice of decision rules. 

Under the vanilla \textsc{DAgger} decision-rule (\Cref{alg:vanilla_dagger_decision_rule}), the expert's action is chosen with probability $\beta_i\in [0,1]$, where $i$ denotes the DAgger epoch. If $\beta_i = \lambda \beta_{i-1}$ for some $\lambda\in(0,1)$, then the novice takes increasingly more actions each epoch. As the novice is given more training labels from previous epochs, it is allowed greater autonomy in exploring the state-space. 

The vanilla \textsc{DAgger} decision-rule does not take into account any similarity measure between the novice and the expert choice of action. Hence, even if the novice suggests a highly unsafe action, vanilla \textsc{DAgger} allows the novice to act with probability $(1-\beta_i)$. The decision-rule employed by \textsc{SafeDAgger}, presented in \Cref{alg:safe_dagger_decision_rule} and referred to as \textsc{SafeDAgger*}, allows the novice to act if the distance between the actions is less than some chosen threshold~$\tau$~\cite{Zhang2016}.\footnote{
	To reduce the number of expert queries, \textsc{SafeDAgger} approximates the \textsc{SafeDAgger*} decision rule via a deep policy that determines whether or not the novice policy is likely to deviate from the reference policy. Unlike \textsc{SafeDAgger}, we are not concerned with minimizing expert queries. Hence, we compare to the \textsc{SafeDAgger*} decision rule directly, as opposed to the approximation.
}

An ideal decision rule would allow the novice to act if there is a sufficiently low probability that the system can transition to an unsafe state. If the combined system is currently near an unsafe state, the tolerable perturbation from the expert's choice of action is smaller than when the system is far from unsafe states. Hence, the single threshold $\tau$ employed in \textsc{SafeDAgger*} is either too conservative when the system is far from unsafe states or too relaxed when near them.

To approximate the ideal decision rule in a model-free manner, we propose considering the distance between the novice's and expert's actions as well as the entropy in the novice policy. To estimate the uncertainty of the novice policy, we utilize Bayesian Deep Learning.

\begin{algorithm}[!t]
	\caption{\textsc{VanillaDAgger} Decision Rule \label{alg:vanilla_dagger_decision_rule}}
	\begin{algorithmic}[1]
		\Procedure{DR}{$o_t, i, \beta_0, \lambda$}
		\State $a_{\text{nov}, t} \gets \pi_{\text{nov},i}(o_t)$
		\State $a_{\text{exp}, t} \gets \pi_{\text{exp}}(o_t)$
		\State $\beta_i \gets \lambda^i\beta_0$
		\State $ z \sim \text{Uniform}(0,1)$
		\If{$z \leq \beta_i$}
		\State \textbf{return} $a_{\text{exp}, t}$
		\Else
		\State \textbf{return} $a_{\text{nov}, t}$
		\EndIf
		\EndProcedure
	\end{algorithmic}
\end{algorithm}

\begin{algorithm}[!t]
	\caption{\textsc{SafeDAgger*} Decision Rule\label{alg:safe_dagger_decision_rule}}
	\begin{algorithmic}[1]
		\Procedure{DR}{$o_t, \tau$}
		\State $a_{\text{nov}, t} \gets \pi_{\text{nov},i}(o_t)$
		\State $a_{\text{exp}, t} \gets \pi_{\text{exp}}(o_t)$
		\If{$\lVert a_{\text{nov}, t} - a_{\text{exp}, t}\rVert \leq \tau$}
		\State \textbf{return} $a_{\text{nov}, t}$
		\Else
		\State \textbf{return} $a_{\text{exp}, t}$
		\EndIf
		\EndProcedure
	\end{algorithmic}
\end{algorithm}

\begin{figure*}
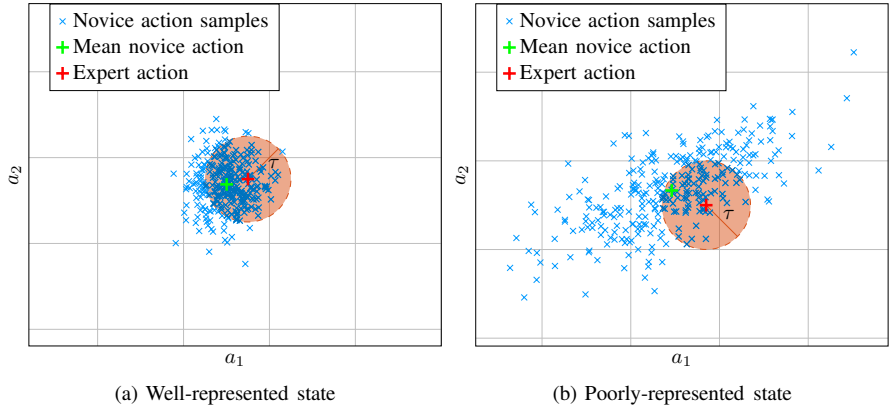

	\centering
	\subfloat[][Well-represented state]{\scalebox{.8}{\input{dist_fig2.tex}}\label{fig:dist_fig_a}}
	\subfloat[][Poorly-represented state]{\scalebox{.8}{\input{dist_fig.tex}}\label{fig:dist_fig_b}}
	\caption{\small Example computation of the \textsc{DropoutDAgger} decision rule governing whether the expert's or novice's action is chosen at a given time-step. The action space is two-dimensional, with action $\vec{a} = [a_1, a_2]$. The novice policy is queried $N$ times to estimate the probability $\hat{p}$ is within a ball of radius $\tau$ centered at the expert action. If $\hat{p} \geq p$, for chosen threshold $p$, then we choose the mean novice action. Otherwise, we choose the expert's action. \Cref{fig:dist_fig_a} shows an example state that is well-represented in $\mathcal{D}$. Because we have many expert labels for this state, the novice policy is low-entropy and centered near the expert's action. Hence, both \textsc{DropoutDAgger} and \textsc{SafeDAgger*} decision rules would allow the novice to act. However, in \Cref{fig:dist_fig_b}, the state is poorly-represented in $\mathcal{D}$, and the novice policy is consequently high-entropy. The \textsc{DropoutDAgger} decision rule would not allow the novice to act if $p$ is appropriately chosen, but the \textsc{SafeDAgger*} decision rule still would allow the novice to act. Vanilla \textsc{DAgger} would choose between the two with a weighted coin-flip, with no regard to the similarity of the actions. 
		\label{fig:dist_fig}}
\end{figure*}

\subsection{Bayesian Approximation via Dropout}

To overcome the fact that deep learning lacks the ability to reason about model uncertainty \cite{oliveira2016known}, Gal et al. have worked towards approximating Bayesian models with neural networks through dropout \cite{Gal2015DropoutB}.
By incorporating dropout at every weight layer in a network, an approximation of a Gaussian process is obtained. 
Given a policy trained with dropout and an input observation, we query the network $N$ times to obtain a distribution over actions, using randomly sampled dropout masks.
For more details, we guide the reader to \cite{Gal2015DropoutB,srivastava2014dropout}.

By invoking dropout, our novice policy approximates a Gaussian process that will produce a low-entropy distribution over actions that is centered around the expert's action, if the input observation is well represented in $\mathcal{D}$. Further, the novice policy will produce high-entropy distributions over actions if the input observation is unlike what has been labeled by the expert in $\mathcal{D}$.

\section{DROPOUT DAGGER}
\label{sec:methods}
We present the \textsc{DropoutDAgger} decision rule, in which we choose the mean action of the novice only if its distribution over actions has sufficient probability mass around the action suggested by the expert. The algorithm, described in \Cref{alg:dropout_decision_rule}, is parameterized by $\tau$, which specifies the size of a ball around the expert's action, and $p$, which is a threshold for the probability mass we desire to be inside this ball, if the novice is allowed to act. An example of computation of this decision rule is shown in \Cref{fig:dist_fig}. We approximate this distribution over actions by first requiring that the neural-network policy is trained dataset $\mathcal{D}$ using dropout, and then querying the network multiple times with the current observation and random dropout masks.

\begin{algorithm}[!t]
	\caption{\textsc{DropoutDAgger} Decision Rule\label{alg:dropout_decision_rule}}
	\begin{algorithmic}[1]
		\Procedure{DR}{$o_t, \tau, p, N$}
		\State $a_{\text{nov}, t, j\in\{1,\ldots,N\}} \gets \pi_{\text{nov},i}(o_t)$ %\Comment{Query $N$ times}
		\State $a_{\text{exp}, t} \gets \pi_{\text{exp}}(o_t)$
		\State $\hat{p} \gets \frac1N \sum_{j=1}^N \mathbf{1}\{\lVert a_{\text{exp},t} - a_{\text{nov},t,j}\rVert \leq \tau\} $
		\If{$\hat{p} \geq p$}
		\State \textbf{return} $\frac1N \sum_{j=1}^N a_{\text{nov},t,j}$
		\Else
		\State \textbf{return} $a_{\text{exp},t}$
		\EndIf
		\EndProcedure
	\end{algorithmic}
\end{algorithm}

As previously stated, an ideal decision rule choose the novice's action in `low-risk states,' and choose the expert's action in `high-risk states.' By using the \textsc{DropoutDAgger} decision rule, we allow the novice to act in familiar states that are well represented in $\mathcal{D}$, but hand control back to the expert when the combined system enters an unfamiliar region of the state-space. 

A comparison between the vanilla \textsc{DAgger}, \textsc{SafeDAgger*}, and \textsc{DropoutDAgger} decision rules can be seen in Figures~\ref{fig:dist_fig_a} and~\ref{fig:dist_fig_b}. Vanilla \textsc{DAgger} leaves the choice of action up to a weighted coin-flip, with no regard to current state of the system. \textsc{SafeDAgger*} is too restrictive in safer, familiar regions of the state-space to sufficiently guarantee safety in unsafe regions. \textsc{DropoutDAgger} is able to utilize the additional information provided by the distribution over the novice's action in order to allow the novice to control the system in familiar parts of the state space, and hand control back to the expert in unfamiliar parts of the state-space.

By appropriately choosing the hyper-parameters $p$ and $\tau$, we satisfy the dual objectives of allowing the novice to act only if its distribution over actions is sufficiently low-entropy, as well as sufficiently close to the expert's. The dropout probability $d$ should be chosen to reflect the epistemic uncertainty, arising from finite demonstration data, and aleatoric uncertainty, arising from the stochastic environment. The probability $d$ should be selected for either by grid-search to minimize loss on test-demonstration data, or optimized for using `Concrete Dropout,' which uses a continuous relaxation of discrete dropout masks~\cite{Gal2017Concrete}. If we set $d$ to zero, \textsc{DropoutDAgger} effectively reduces to \textsc{SafeDAgger*}. If we set $\tau$ to zero, the algorithm reduces to behavior cloning. If we set $p$ to zero, the algorithm reduces to one in which the expert merely labels the data, but does not ever influence the system during an episode.

\section{EXPERIMENTS}
\label{sec:experiments}

We demonstrate that \textsc{DropoutDAgger} is able to achieve expert-level performance, while maintaining safety during training in two experimental domains. 
For each episode, we use average total reward of the combined expert-novice system as a measure of `safety performance,' and the average total reward of the novice alone as a measure of `learning performance.' An algorithm `safe' if it demonstrates safety performance on par with that of \textsc{BehaviorCloning}, in which only the expert acts. Learning performance is assessed by the rate at which the novice achieves expert-level performance. 

\subsection{MuJoCo HalfCheetah}
\label{sec:half_cheetah}

\begin{figure*}
	\centering
	\scalebox{0.7}{\begin{tikzpicture}[]
\begin{groupplot}[group style={horizontal sep = 1 cm, group size=2 by 1}]
\nextgroupplot [ylabel = {Avg. Total Return}, title = {Safety Performance}, xlabel = {DAgger Epoch}, grid=both]\addplot+ [mark = {}, C1]coordinates {
(0.0, 405.172442003)
(1.0, 397.985495659)
(2.0, 399.523726071)
(3.0, 397.154140504)
(4.0, 399.1624026)
(5.0, 398.67293165)
(6.0, 402.740086915)
(7.0, 396.877740743)
(8.0, 398.685937093)
(9.0, 398.196952793)
};
\addplot+ [mark = {}, C2]coordinates {
(0.0, 400.861930916)
(1.0, 311.522471233)
(2.0, 312.427214694)
(3.0, 278.390786619)
(4.0, 317.378188703)
(5.0, 291.725946644)
(6.0, 331.828509158)
(7.0, 356.967315591)
(8.0, 378.926467331)
(9.0, 373.764283946)
};
\addplot+ [mark = {}, C3]coordinates {
(0.0, 391.11632137)
(1.0, 372.007669245)
(2.0, 371.206392008)
(3.0, 356.793620902)
(4.0, 349.698894596)
(5.0, 362.610686137)
(6.0, 369.553422352)
(7.0, 382.792268589)
(8.0, 380.698614848)
(9.0, 380.973272566)
};
\addplot+ [mark = {}, C4]coordinates {
(0.0, 404.500668473)
(1.0, 400.692117768)
(2.0, 403.065449883)
(3.0, 397.011224744)
(4.0, 396.180873247)
(5.0, 402.891651709)
(6.0, 398.909033946)
(7.0, 392.395724143)
(8.0, 397.453806211)
(9.0, 399.579596028)
};
\addplot+ [mark = {}, C5]coordinates {
(0.0, -20.050831986)
(1.0, -35.0364881324)
(2.0, 105.591322065)
(3.0, 261.788590871)
(4.0, 312.652860973)
(5.0, 347.054653655)
(6.0, 347.252563578)
(7.0, 323.193502097)
(8.0, 367.128400797)
(9.0, 370.274021368)
};
\nextgroupplot [legend style = {{at={(1.05,1.0)},anchor=north west}}, title = {Learning Performance}, xlabel = {DAgger Epoch}, grid=both]\addplot+ [mark = {}, C1]coordinates {
(0.0, 153.797624817)
(1.0, 228.536227448)
(2.0, 258.188336791)
(3.0, 252.153864987)
(4.0, 311.85015192)
(5.0, 331.838053199)
(6.0, 333.701272096)
(7.0, 329.628090843)
(8.0, 362.258479389)
(9.0, 361.619133979)
};
\addlegendentry{DropoutDAgger}
\addplot+ [mark = {}, C2]coordinates {
(0.0, 78.8764589935)
(1.0, 175.566203466)
(2.0, 224.454217636)
(3.0, 285.966627824)
(4.0, 303.672234009)
(5.0, 325.569382954)
(6.0, 360.497545174)
(7.0, 359.583403423)
(8.0, 363.515073244)
(9.0, 376.534642626)
};
\addlegendentry{VanillaDAgger}
\addplot+ [mark = {}, C3]coordinates {
(0.0, 81.3574578475)
(1.0, 94.6921230613)
(2.0, 204.615922825)
(3.0, 202.192957646)
(4.0, 296.05626272)
(5.0, 314.187469408)
(6.0, 362.932136657)
(7.0, 356.677262215)
(8.0, 363.756764776)
(9.0, 375.866846688)
};
\addlegendentry{SafeDAgger*}
\addplot+ [mark = {}, C4]coordinates {
(0.0, 77.9647484586)
(1.0, 133.962622429)
(2.0, 132.81976489)
(3.0, 135.543922144)
(4.0, 227.210648493)
(5.0, 241.405589339)
(6.0, 304.666162989)
(7.0, 306.931914058)
(8.0, 316.713308807)
(9.0, 337.916512817)
};
\addlegendentry{Behavior Cloning}
\addplot+ [mark = {}, C5]coordinates {
(0.0, -40.9552418197)
(1.0, 114.19292053)
(2.0, 263.041959126)
(3.0, 315.091573737)
(4.0, 343.404206599)
(5.0, 340.637760365)
(6.0, 330.345737988)
(7.0, 370.238011979)
(8.0, 372.011873155)
(9.0, 373.969738471)
};
\addlegendentry{ExpertLabelsOnly}
\end{groupplot}

\end{tikzpicture}}

	\caption{\small Performance of variants of the \textsc{DAgger} algorithm on the vanilla MuJoCo HalfCheetah-v1 environment, averaged over 50 episodes. As we can see, \textsc{DropoutDAgger} makes no compromise to the safety of the combined system, while novice appears to learn a well-performing policy as quickly as other algorithms. \label{fig:hcnsn}}
\end{figure*}
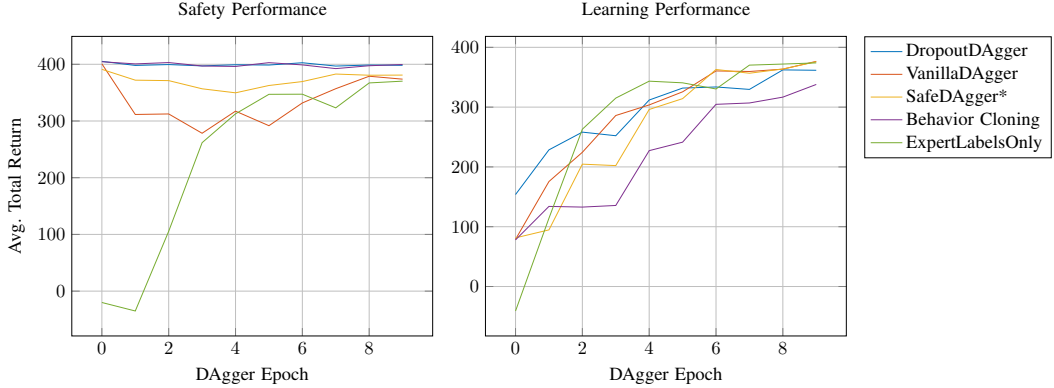

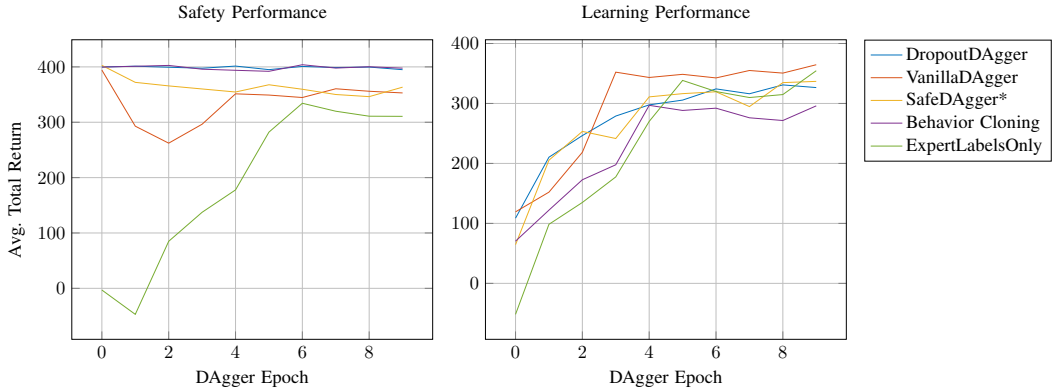
\begin{figure*}
	\centering
	\scalebox{0.7}{\begin{tikzpicture}[]
\begin{groupplot}[group style={horizontal sep = 1 cm, group size=2 by 1}]
\nextgroupplot [ylabel = {Avg. Total Return}, title = {Safety Performance}, xlabel = {DAgger Epoch}, grid=both]\addplot+ [mark = {}, C1]coordinates {
(0.0, 399.802763435)
(1.0, 401.238974094)
(2.0, 399.393333107)
(3.0, 397.578814352)
(4.0, 401.458407303)
(5.0, 395.042909105)
(6.0, 400.813568101)
(7.0, 398.8155533)
(8.0, 399.6895785)
(9.0, 394.963188974)
};
\addplot+ [mark = {}, C2]coordinates {
(0.0, 394.20223309)
(1.0, 292.912840092)
(2.0, 262.234771445)
(3.0, 296.441521756)
(4.0, 351.330280409)
(5.0, 349.114339888)
(6.0, 344.684109478)
(7.0, 360.475404592)
(8.0, 355.969472955)
(9.0, 352.958567654)
};
\addplot+ [mark = {}, C3]coordinates {
(0.0, 402.766400362)
(1.0, 372.065040209)
(2.0, 365.667746523)
(3.0, 360.125980879)
(4.0, 354.654214307)
(5.0, 367.727953057)
(6.0, 359.606001877)
(7.0, 350.119704586)
(8.0, 346.266782437)
(9.0, 363.397067251)
};
\addplot+ [mark = {}, C4]coordinates {
(0.0, 399.659874605)
(1.0, 401.111154979)
(2.0, 402.668725262)
(3.0, 395.825618428)
(4.0, 393.843085537)
(5.0, 392.004745371)
(6.0, 403.915425525)
(7.0, 397.856169061)
(8.0, 400.272349382)
(9.0, 397.689471022)
};
\addplot+ [mark = {}, C5]coordinates {
(0.0, -3.02583240206)
(1.0, -46.9976292782)
(2.0, 84.9809558049)
(3.0, 137.383002031)
(4.0, 177.79846887)
(5.0, 282.160231076)
(6.0, 334.296620753)
(7.0, 319.918731678)
(8.0, 310.920057427)
(9.0, 310.656561336)
};
\nextgroupplot [legend style = {{at={(1.05,1.0)},anchor=north west}}, title = {Learning Performance}, xlabel = {DAgger Epoch}, grid=both]\addplot+ [mark = {}, C1]coordinates {
(0.0, 108.713758017)
(1.0, 210.390809591)
(2.0, 246.458454521)
(3.0, 279.012468578)
(4.0, 297.644629761)
(5.0, 305.632232556)
(6.0, 324.305774741)
(7.0, 316.1815422)
(8.0, 330.942796371)
(9.0, 326.508766608)
};
\addlegendentry{DropoutDAgger}
\addplot+ [mark = {}, C2]coordinates {
(0.0, 119.247620835)
(1.0, 152.110398143)
(2.0, 218.556908653)
(3.0, 352.210259019)
(4.0, 343.333361686)
(5.0, 348.574091287)
(6.0, 342.585210778)
(7.0, 354.978363528)
(8.0, 350.637509934)
(9.0, 364.602809176)
};
\addlegendentry{VanillaDAgger}
\addplot+ [mark = {}, C3]coordinates {
(0.0, 64.4996764676)
(1.0, 205.417756295)
(2.0, 253.378857364)
(3.0, 241.56990447)
(4.0, 310.982215116)
(5.0, 316.362601702)
(6.0, 319.512335334)
(7.0, 294.796230236)
(8.0, 334.752298468)
(9.0, 336.844190518)
};
\addlegendentry{SafeDAgger*}
\addplot+ [mark = {}, C4]coordinates {
(0.0, 70.3646280693)
(1.0, 122.118069065)
(2.0, 172.780670358)
(3.0, 197.767579018)
(4.0, 296.880915553)
(5.0, 288.351802698)
(6.0, 292.141596504)
(7.0, 276.106399624)
(8.0, 271.570771254)
(9.0, 296.007356762)
};
\addlegendentry{Behavior Cloning}
\addplot+ [mark = {}, C5]coordinates {
(0.0, -51.6642883288)
(1.0, 98.4670871443)
(2.0, 134.912324758)
(3.0, 177.516333548)
(4.0, 270.053271759)
(5.0, 338.410427863)
(6.0, 319.905735938)
(7.0, 309.887294165)
(8.0, 314.84055446)
(9.0, 354.604275445)
};
\addlegendentry{ExpertLabelsOnly}
\end{groupplot}

\end{tikzpicture}}
	\caption{\small Performance of variants of the \textsc{DAgger} algorithm on a variant of MuJoCo HalfCheetah-v1 environment with noisy observations, averaged over 50 episodes. Increased uncertainty compromises the safety of all variants of \textsc{DAgger} except \textsc{DropoutDAgger} and \textsc{BehaviorCloning}. While \textsc{DropoutDAgger} continues to provide safety, the learning performance remains comparable to other algorithms. \label{fig:hcsnn}}
	\vspace{-.1in}
\end{figure*}

\begin{table}
	\centering
	\caption{Hyperparameters used to train Expert Policy for Half-Cheetah domain.}
	\label{tab:halfcheetahTRPO}
	\begin{tabular}{@{}llr@{}} \toprule
		Parameter & Value & Unit \\ \midrule
		Algorithm & TRPO & \\
		MLP Hidden Layer Sizes & $(64, 64)$ & neurons \\
		$\gamma$ & $0.99$ & \\
		$\lambda$ & $0.97$ & \\
		TRPO Max Step & 0.01 & \\
		Batch Size & 25000 & $\frac{\text{timesteps}}{\text{epoch}}$ \\
		Max. Episode Length & 100 & timesteps \\
		Environment Seed & 1 & \\
		\bottomrule
	\end{tabular}
\end{table}

The MuJoCo HalfCheetah-v1 domain is an OpenAI Gym Environment with observations in $\mathbb{R}^{20}$ and actions in $\mathbb{R}^6$~\cite{openaigym}. The environment provides reward proportional to the horizontal distance traveled. 
An optimal policy propels the half-cheetah robot into a steady run, going as far forward as possible in the time it has. 
First, we train a multi layer perceptron (MLP) policy to act as the expert, and then compare the performance of \textsc{DropoutDAgger} to other \textsc{DAgger} variants. 

We compare two scenarios. First, the novice is given the same observation as the expert. Second, the novice sees the observation corrupted by diagonal-Gaussian noise with $\sigma=0.1$, representing settings where the expert and novice see different observations. The added noise increases aleatoric uncertainty, which degrades performance of naive imitation learning approaches.

The optimal policy is trained using the TRPO hyperparameters summarized in \Cref{tab:halfcheetahTRPO}~\cite{islam2017reproducibility}, using the \texttt{rllab} implementation of TRPO~\cite{rllab,schulman2015trust}.
The MLP representing the novice policy has two hidden layers with 64 hidden units each, followed by a hidden layer with 32 hidden units. When training the neural network on a given dataset $\mathcal{D}$, an ADAM optimizer is used with a learning rate of $10^{-3}$, $\beta_1 = 0.9$, and $\beta_2 = 0.999$. Weights are $l_2$ regularized with regularization weight $10^{-5}$. The \textsc{DropoutDAgger} policy is trained with a dropout probability $d$ of $0.05$. The network is trained for 100 epochs, with a mini-batch size of 32. 

We compare \textsc{DropoutDAgger} to \textsc{BehaviorCloning}, in which the decision-rule always chooses the expert's action, \textsc{ExpertLabelsOnly}, in which the decision-rule always chooses the novice's action, Vanilla \textsc{DAgger}, and \textsc{SafeDAgger*}. When testing Vanilla \textsc{DAgger}, we use $\beta_0 = 1$ and $\beta_i = 0.63 \beta_{i-1}$, which brings $\beta$ down to 0.01 by the tenth DAgger epoch. Hyperparameters of $\tau=0.3$ and $p = 0.6$ are used for \textsc{DropoutDAgger}, and $\tau=0.6$ is used for \textsc{SafeDAgger*}, chosen by grid search. 

The performance of each policy on the environment are averaged over 50 episodes to estimate of safety and learning performance.
Figures~\ref{fig:hcnsn} and~\ref{fig:hcsnn} show that \textsc{DropoutDAgger} does not compromise the safety of the combined expert-novice system, while being able to train a well-performing novice policy at a rate comparable to other variants of DAgger. 

Since \textsc{BehaviorCloning} never chooses the novice action, it unsurprisingly perfectly safe. However, we see that all other algorithms except \textsc{DropoutDAgger} compromise the safety. Since the dataset generated by \textsc{BehaviorCloning} contains autocorrelated samples drawn only from nominal expert trajectories, it has poor learning performance. \textsc{DropoutDAgger} both maintains safety performance and achieves a learning performance comparable to all other \textsc{DAgger} variants. 

We see in \Cref{fig:hcsnn} that adding observation noise adversely affects the learning performance of all algorithms. This consequently adversely affects the safety performance of Vanilla \textsc{DAgger}, \textsc{SafeDAgger*}, and of course \textsc{ExpertLabelsOnly}, but does not compromise the safety performance of \textsc{DropoutDAgger}. 

\subsection{Dubins Car Lidar}

\begin{figure}
	\centering
	\scalebox{0.8}{\input{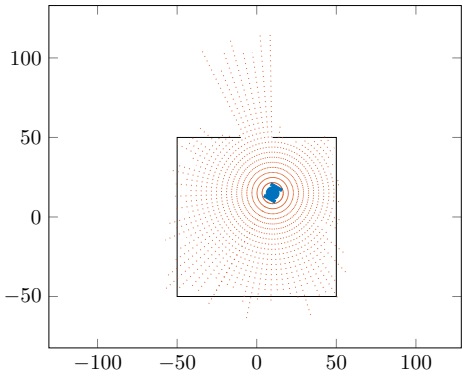}}
	\caption{\small The Dubins Car Lidar environment. \label{fig:dubinsenv}}
	\vspace{-.1in}
\end{figure}

\begin{table}[!t]
\centering
\caption{Dubins Car Lidar Parameters}
\label{tab:dubcarenvparams}
\begin{tabular}{@{}llr@{}} \toprule
Parameter & Value & Unit \\ \midrule
Room Height/Width & 100 & \si{\meter} \\
Exit Width & 20 & \si{\meter} \\
Lidar Resolution & 100 & rays \\
Lidar Max. Range & 100 & \si{\meter} \\
$\sigma_1$ & 10 & \si{\meter} \\
$\sigma_2$ & 10 & $\frac{\si{\meter}}{\si{\meter}}$ \\
Timestep & 0.1 & \si{\second} \\
Max. Angular Veloicty & $\pm1.0$ & $\frac{\si{\radian}}{\si{\second}}$ \\
\bottomrule
\end{tabular}
\end{table}

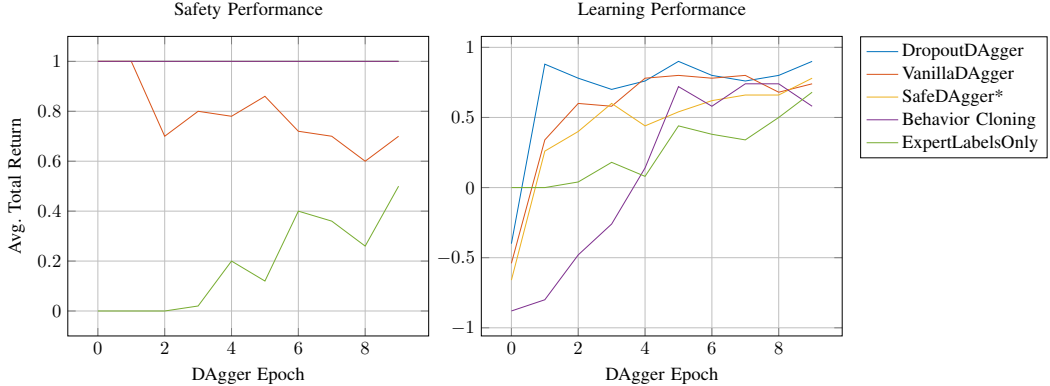
\begin{figure*}
	\centering
	\scalebox{0.7}{\begin{tikzpicture}[]
\begin{groupplot}[group style={horizontal sep = 1 cm, group size=2 by 1}]
\nextgroupplot [ylabel = {Avg. Total Return}, title = {Safety Performance}, xlabel = {DAgger Epoch}, grid=both]\addplot+ [mark = {}, C1]coordinates {
(0.0, 1.0)
(1.0, 1.0)
(2.0, 1.0)
(3.0, 1.0)
(4.0, 1.0)
(5.0, 1.0)
(6.0, 1.0)
(7.0, 1.0)
(8.0, 1.0)
(9.0, 1.0)
};
\addplot+ [mark = {}, C2]coordinates {
(0.0, 1.0)
(1.0, 1.0)
(2.0, 0.7)
(3.0, 0.8)
(4.0, 0.78)
(5.0, 0.86)
(6.0, 0.72)
(7.0, 0.7)
(8.0, 0.6)
(9.0, 0.7)
};
\addplot+ [mark = {}, C3]coordinates {
(0.0, 1.0)
(1.0, 1.0)
(2.0, 1.0)
(3.0, 1.0)
(4.0, 1.0)
(5.0, 1.0)
(6.0, 1.0)
(7.0, 1.0)
(8.0, 1.0)
(9.0, 1.0)
};
\addplot+ [mark = {}, C4]coordinates {
(0.0, 1.0)
(1.0, 1.0)
(2.0, 1.0)
(3.0, 1.0)
(4.0, 1.0)
(5.0, 1.0)
(6.0, 1.0)
(7.0, 1.0)
(8.0, 1.0)
(9.0, 1.0)
};
\addplot+ [mark = {}, C5]coordinates {
(0.0, 0.0)
(1.0, 0.0)
(2.0, 0.0)
(3.0, 0.02)
(4.0, 0.2)
(5.0, 0.12)
(6.0, 0.4)
(7.0, 0.36)
(8.0, 0.26)
(9.0, 0.5)
};
\nextgroupplot [legend style = {{at={(1.05,1.0)},anchor=north west}}, title = {Learning Performance}, xlabel = {DAgger Epoch}, grid=both]\addplot+ [mark = {}, C1]coordinates {
(0.0, -0.4)
(1.0, 0.88)
(2.0, 0.78)
(3.0, 0.7)
(4.0, 0.76)
(5.0, 0.9)
(6.0, 0.8)
(7.0, 0.76)
(8.0, 0.8)
(9.0, 0.9)
};
\addlegendentry{DropoutDAgger}
\addplot+ [mark = {}, C2]coordinates {
(0.0, -0.54)
(1.0, 0.34)
(2.0, 0.6)
(3.0, 0.58)
(4.0, 0.78)
(5.0, 0.8)
(6.0, 0.78)
(7.0, 0.8)
(8.0, 0.68)
(9.0, 0.74)
};
\addlegendentry{VanillaDAgger}
\addplot+ [mark = {}, C3]coordinates {
(0.0, -0.66)
(1.0, 0.26)
(2.0, 0.4)
(3.0, 0.6)
(4.0, 0.44)
(5.0, 0.54)
(6.0, 0.62)
(7.0, 0.66)
(8.0, 0.66)
(9.0, 0.78)
};
\addlegendentry{SafeDAgger*}
\addplot+ [mark = {}, C4]coordinates {
(0.0, -0.88)
(1.0, -0.8)
(2.0, -0.48)
(3.0, -0.26)
(4.0, 0.14)
(5.0, 0.72)
(6.0, 0.58)
(7.0, 0.74)
(8.0, 0.74)
(9.0, 0.58)
};
\addlegendentry{Behavior Cloning}
\addplot+ [mark = {}, C5]coordinates {
(0.0, 0.0)
(1.0, 0.0)
(2.0, 0.04)
(3.0, 0.18)
(4.0, 0.08)
(5.0, 0.44)
(6.0, 0.38)
(7.0, 0.34)
(8.0, 0.5)
(9.0, 0.68)
};
\addlegendentry{ExpertLabelsOnly}
\end{groupplot}

\end{tikzpicture}}
	\caption{\small Performance of variants of the \textsc{DAgger} algorithm on the Dubins Car Lidar environment, averaged over 50 episodes. Even with high aleatoric uncertainty, \textsc{DropoutDAgger} has best learning and safety performance. \textsc{BehaviorCloning} and \textsc{SafeDAgger*} do not compromise safety, but exhibit worse learning performance.  \label{fig:dubinsalgocomp}} \end{figure*}

\begin{figure*}
	\centering
	\scalebox{0.7}{\begin{tikzpicture}[]
\begin{axis}[legend style = {{at={(1.05,1.0)},anchor=north west}}, ylabel = {Avg. Total Return}, xlabel = {DAgger Epoch}, grid=both]\addplot+ [mark = {}, solid, C1]coordinates {
(0.0, -0.92)
(1.0, -0.96)
(2.0, -0.68)
(3.0, -0.76)
(4.0, 0.38)
(5.0, 0.5)
(6.0, 0.26)
(7.0, 0.5)
(8.0, 0.6)
(9.0, 0.6)
};
\addlegendentry{$\tau=0.1, p=0.3, d=0.05$}
\addplot+ [mark = {}, solid, C2]coordinates {
(0.0, -0.8)
(1.0, 0.22)
(2.0, 0.3)
(3.0, 0.24)
(4.0, 0.54)
(5.0, 0.74)
(6.0, 0.8)
(7.0, 0.66)
(8.0, 0.66)
(9.0, 0.68)
};
\addlegendentry{$\tau=0.1, p=0.6, d=0.05$}
\addplot+ [mark = {}, solid, C3]coordinates {
(0.0, 0.28)
(1.0, 0.56)
(2.0, 0.6)
(3.0, 0.6)
(4.0, 0.72)
(5.0, 0.72)
(6.0, 0.74)
(7.0, 0.8)
(8.0, 0.62)
(9.0, 0.82)
};
\addlegendentry{$\tau=0.3, p=0.3, d=0.05$}
\addplot+ [mark = {}, solid, C4]coordinates {
(0.0, -0.4)
(1.0, 0.88)
(2.0, 0.78)
(3.0, 0.7)
(4.0, 0.76)
(5.0, 0.9)
(6.0, 0.8)
(7.0, 0.76)
(8.0, 0.8)
(9.0, 0.9)
};
\addlegendentry{$\tau=0.3, p=0.6, d=0.05$}
\addplot+ [mark = {}, solid, C5]coordinates {
(0.0, -0.78)
(1.0, 0.2)
(2.0, 0.28)
(3.0, 0.54)
(4.0, 0.72)
(5.0, 0.8)
(6.0, 0.72)
(7.0, 0.7)
(8.0, 0.68)
(9.0, 0.4)
};
\addlegendentry{$\tau=0.6, p=0.6, d=0.05$}
\addplot+ [mark = {}, dashed, C1]coordinates {
(0.0, -0.8)
(1.0, -0.34)
(2.0, 0.54)
(3.0, 0.48)
(4.0, 0.68)
(5.0, 0.38)
(6.0, 0.6)
(7.0, 0.72)
(8.0, 0.64)
(9.0, 0.78)
};
\addlegendentry{$\tau=0.1, p=0.3, d=0.1$}
\addplot+ [mark = {}, dashed, C2]coordinates {
(0.0, 0.44)
(1.0, 0.44)
(2.0, 0.52)
(3.0, 0.52)
(4.0, 0.44)
(5.0, 0.6)
(6.0, 0.82)
(7.0, 0.8)
(8.0, 0.78)
(9.0, 0.82)
};
\addlegendentry{$\tau=0.1, p=0.6, d=0.1$}
\addplot+ [mark = {}, dashed, C3]coordinates {
(0.0, -0.44)
(1.0, 0.34)
(2.0, 0.52)
(3.0, 0.5)
(4.0, 0.8)
(5.0, 0.52)
(6.0, 0.82)
(7.0, 0.82)
(8.0, 0.66)
(9.0, 0.68)
};
\addlegendentry{$\tau=0.3, p=0.3, d=0.1$}
\addplot+ [mark = {}, dashed, C4]coordinates {
(0.0, -0.3)
(1.0, 0.32)
(2.0, 0.58)
(3.0, 0.68)
(4.0, 0.74)
(5.0, 0.76)
(6.0, 0.54)
(7.0, 0.52)
(8.0, 0.74)
(9.0, 0.82)
};
\addlegendentry{$\tau=0.3, p=0.6, d=0.1$}
\addplot+ [mark = {}, dashed, C5]coordinates {
(0.0, -0.5)
(1.0, -0.2)
(2.0, 0.06)
(3.0, 0.54)
(4.0, 0.58)
(5.0, 0.66)
(6.0, 0.6)
(7.0, 0.44)
(8.0, 0.68)
(9.0, 0.34)
};
\addlegendentry{$\tau=0.6, p=0.6, d=0.1$}
\end{axis}

\end{tikzpicture}}
	\caption{\small Comparison of learning performance of various hyperparameters used for the \textsc{DropoutDAgger} algorithm.  \label{fig:dubinsparamcomp}} 
\end{figure*}
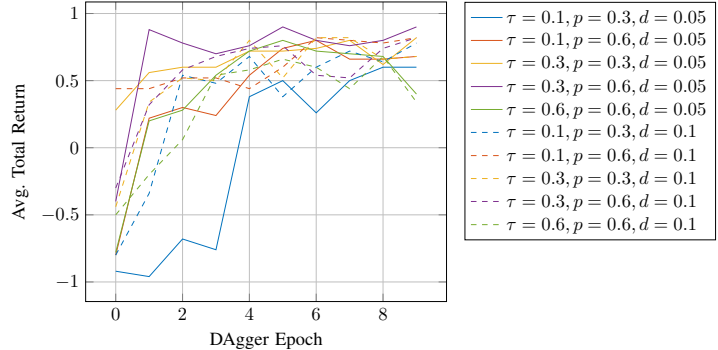

In the Dubins Car Lidar domain, we demonstrate that \textsc{DropoutDAgger} can safely learn a policy in spite of high aleatoric uncertainty. This environment, depicted in \Cref{fig:dubinsenv}, consists of a simple Dubins car that navigates out of a room. 
A \textit{Dubins path} is represented by a circular arc of radius $R$, a straight path, and a second circular arc of radius $R$. 
Given any two poses ($x,y,\theta$) sufficiently far apart with some achievable turning radius, Dubins path can take a Dubins car from the first pose to the second. A finite-state controller acts as the expert, following a Dubins path at a fixed velocity from the initial state to goal state pointing out of the room in the exit.

The expert policy is given access to its exact pose, but the novice policy only has access to noisy `lidar' measurements. These are range measurements to the nearest obstacle along 100 equally spaced rays propagating from the center of the Dubins car (\Cref{fig:dubinsenv}). 
The following noise model gives the corrupted measurement $\hat{x}$:
\begin{align}
	\tilde{x} &= z_1 + (1+z_2)x \\
	\hat{x} &= \max\left(\min(\tilde{x},L),0\right)
\end{align}
where $x$ is the original measurement, $z_i \sim \mathcal{N}(0, \sigma_i)$ for $i \in \{1,2\}$, and $L$ is the maximum lidar range.

We train an MLP to map the lidar measurements to the car's angular velocity.
The environment parameters are summarized in \Cref{tab:dubcarenvparams}. 
The algorithms and optimizer parameters are identical to those in Section \ref{sec:half_cheetah}. 

In \Cref{fig:dubinsalgocomp}, we see that \textsc{DropoutDAgger} maintains perfect safety and good learning performance. Under high aleatoric uncertainty, the \textsc{BehaviorCloning} learning performance deteriorates, thus highlighting the importance of exploration for robustness.

\Cref{fig:dubinsparamcomp} shows the learning performance of various choices of hyperparameters for \textsc{DropoutDAgger}. Though all variants enjoy perfect safety performance, we observe that reducing $\tau$ or increasing $p$ make the algorithm more conservative and reduce learning performance to that of \textsc{BehaviorCloning}, as expected. It is interesting to note that increasing the dropout probability $d$ from 0.05 to 0.1 appears to reduce the sensitivity of the learning performance to the choice of $\tau$ and $p$. 

\section{DISCUSSION}
\label{sec:conclusion}

Naive algorithms like \textsc{BehaviorCloning} rely on a large set of demonstrations to provide a good dataset for learning.
\textsc{DropoutDAgger} extends naive imitation learning, adding the ability to safely explore the state-space. 
Using the novice's action distribution, the \textsc{DropoutDAgger} decision rule allows the novice to act when in familiar regions of the state-space, but returns control to the expert when entering unfamiliar regions. 
Our experiments show that \textsc{DropoutDAgger} allows the combined system to safely gather data and explore poorly represented states. \textsc{DropoutDAgger} demonstrates no compromise to safety and learning performance comparable to other algorithms.

Though \textsc{DropoutDAgger} exhibits a good mix of safety and learning performance, the algorithm still depends on three hyperparameters. In particular, we observe that the choice of dropout probability $d$ can affect the sensitivity of learning performance to these parameters.
Future work includes exploring the optimization of the dropout mask using Concrete Dropout at every epoch~\cite{Gal2017Concrete}.
Additionally, since all the methods described here only apply to continuous actions, we hope to extend the presented tools to discrete and hybrid action spaces.

%\small{
  \bibliographystyle{IEEEtran}
%  \bibliography{references.bib}
  \bibliography{references}
%}

\end{document}